\documentclass{bmvc2k}

\title{SciPostLayout: A Dataset for Layout Analysis and Layout Generation of Scientific Posters}

\addauthor{Shohei Tanaka}{shohei.tanaka@sinicx.com}{1}
\addauthor{Hao Wang}{conan1024hao@akane.waseda.jp}{2,1}
\addauthor{Yoshitaka Ushiku}{yoshitaka.ushiku@sinicx.com}{1}

\addinstitution{
 OMRON SINIC X Corporation\\
 Nagase Hongo Building 3F, 5-24-5 Hongo,\\
 Bunkyo-ku, Tokyo, Japan
}
\addinstitution{
 Waseda University\\
 1-6-1 Nishi-Waseda, Shinjuku-ku,\\
 Tokyo, Japan
}

\runninghead{Shohei Tanaka, Hao Wang, Yoshitaka Ushiku}{SciPostLayout}

\begin{document}

\maketitle

\begin{abstract}
Scientific posters are used to present the contributions of scientific papers effectively in a graphical format. However, creating a well-designed poster that efficiently summarizes the core of a paper is both labor-intensive and time-consuming. A system that can automatically generate well-designed posters from scientific papers would reduce the workload of authors and help readers understand the outline of the paper visually. Despite the demand for poster generation systems, only a limited research has been conduced due to the lack of publicly available datasets. Thus, in this study, we built the SciPostLayout dataset, which consists of 7,855 scientific posters and manual layout annotations for layout analysis and generation. SciPostLayout also contains 100 scientific papers paired with the posters. All of the posters and papers in our dataset are under the CC-BY license and are publicly available. As benchmark tests for the collected dataset, we conducted experiments for layout analysis and generation utilizing existing computer vision models and found that both layout analysis and generation of posters using SciPostLayout are more challenging than with scientific papers. We also conducted experiments on generating layouts from scientific papers to demonstrate the potential of utilizing LLM as a scientific poster generation system. The dataset is publicly available at \url{https://huggingface.co/datasets/omron-sinicx/scipostlayout_v2}. The code is also publicly available at \url{https://github.com/omron-sinicx/scipostlayout}.
\end{abstract}

\section{Introduction}

\begin{figure}[t]
\centering
\includegraphics[width=0.8\linewidth]{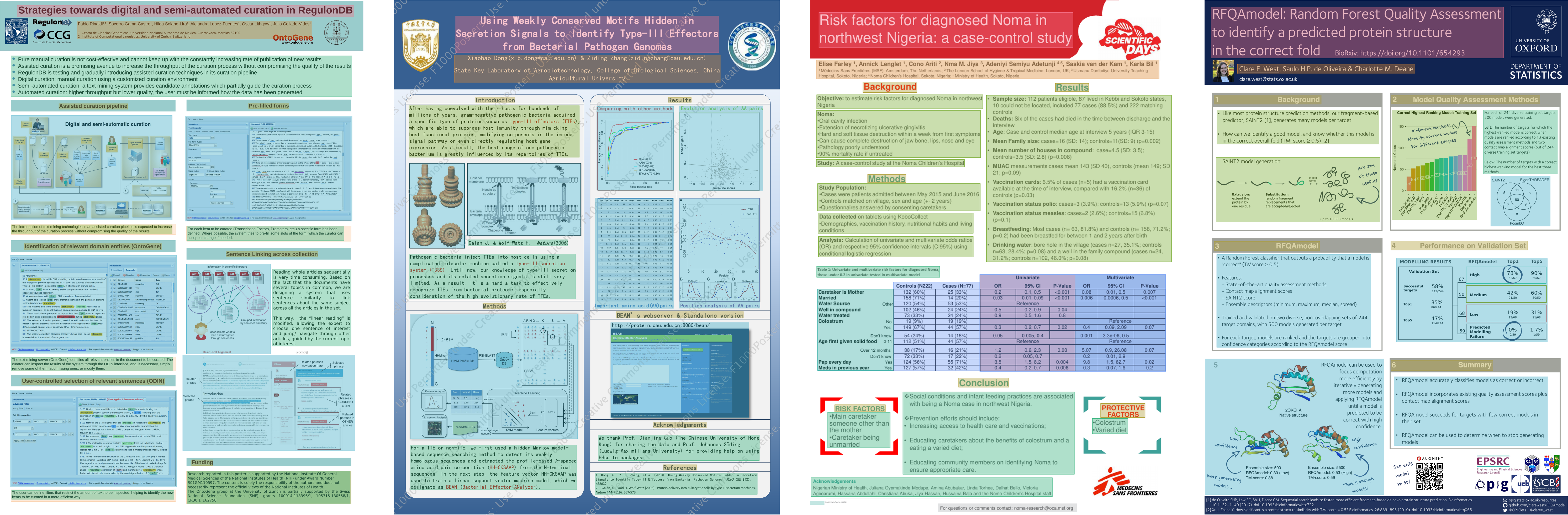}
\vspace{-2mm}
\caption{Example posters and annotations in SciPostLayout.}
\label{fig:example_posters}
\end{figure}

Scientific posters are used to efficiently present the contributions of a scientific paper in a graphical format.
A well-designed scientific poster conveys the essential elements of the research and requires less time to read than a paper.
Unfortunately, creating a scientific poster that efficiently summarizes a paper is both labor-intensive and time-consuming.
Although automating this task by utilizing ML models has shown promise, research on scientific poster generation remains scarce due to the high complexity and multimodality of the task.
Previous studies~\cite{1547260908565-217496438,xu2021neural} have built datasets to evaluate poster generation systems for scientific papers, but these datasets are either not publicly available or the data license is unclear, leaving this research field lacking in gold-standard benchmarks.
  
Previous studies~\cite{zhong2019publaynet,deka2017rico} have published publicly available datasets for layout analysis~\cite{gu2021unidoc,huang2022layoutlmv3,li2022dit,zhong2019publaynet} and layout generation~\cite{deka2017rico,fu2022doc2ppt,https://doi.org/10.1111/cgf.14399,yamaguchi2021canvasvae,zheng2019content}.
However, these datasets focus on scientific papers or mobile application design, and neglect to examine the layout of scientific poster.
Layout analysis and layout generation are important tasks for identifying the characteristics of effective layout design in a particular domain.
Developing layout analysis models would be helpful in terms of identifying layout patterns common to well-designed posters.
In addition, by developing layout generation models, we can generate layouts that follow the revealed patterns of attractive posters.
In other words, building a layout dataset of scientific posters that can be utilized for these tasks would contribute significantly to the research on scientific poster generation.

In this paper, we introduce SciPostLayout, the first scientific poster layout dataset for layout analysis and generation.
This dataset includes 7,855 scientific posters with manual layout annotations.
All posters included in the dataset are under the CC-BY license~\footnote{https://creativecommons.org/licenses/by/4.0/. Note that some of the older posters are under older licenses, such as CC-BY 2.0 and CC-BY 3.0.}.
Figure~\ref{fig:example_posters} shows examples of posters and annotations of SciPostLayout. 
SciPostLayout can be used to evaluate layout analysis and generation systems in the same way as existing datasets.
Both the layout analysis and generation on SciPostLayout are more challenging than scientific papers because of the various positions of elements such as figures and tables.
In addition, we manually collected 100 papers associated with posters in the dataset to utilize SciPostLayout in layout generation from scientific papers.
These papers are also under the CC-BY license.

We evaluated existing models for layout analysis and generation on SciPostLayout.
In the layout analysis task, we found that although the existing models can recognize certain elements with high accuracy, they are less effective than on the scientific paper dataset, indicating that layout analysis in SciPostLayout is more complicated.
In the layout generation task, an LLM-based model can generate aligned layouts with less overlap, but we found that layout generation on SciPostLayout is also a more challenging task compared to generation on the scientific paper dataset.
In addition, we implemented LLM-based models for generating poster layouts from papers and discussed the potential of poster generation utilizing LLM.


\section{Related Work}

In this section, we describe related work on layout analysis, layout generation, and scientific poster generation.

\subsection{Layout Analysis}

Layout analysis is a task in which a system detects the layouts of unstructured documents by predicting bounding boxes and categories such as tables or figures.
This task is generally categorized as an object detection problem~\cite{gu2021unidoc,li2022dit,zhong2019publaynet}.

The PubLayNet dataset~\cite{zhong2019publaynet} is one of the well-known layout analysis datasets.
It contains scientific paper images automatically annotated with bounding boxes and polygonal segmentation across five categories: text, title, list, figure, and table.
Previous studies have demonstrated that deep neural networks, including UDoc~\cite{gu2021unidoc}, DiT~\cite{li2022dit}, and LayoutLMv3~\cite{huang2022layoutlmv3}, which are trained on the PubLayNet dataset, can recognize the layouts of scientific papers with high accuracy.
These models utilize Faster R-CNN~\cite{NIPS201514bfa6bb}, Mask R-CNN~\cite{he2017mask}, and Cascade R-CNN~\cite{cai2018cascade} as object detection models and ResNet~\cite{he2016deep} and Vision Transformer~\cite{dosovitskiy2020image} as visual backbones.

Other studies have also proposed layout analysis datasets, such as Article Regions~\cite{soto-yoo-2019-visual}, which is a dataset of region-annotated scientific papers from PubMed Central, TableBank~\cite{li2019tablebank}, an image-based table detection dataset built with Word and LaTeX documents, and DocBank~\cite{li-etal-2020-docbank}, a document-level dataset with token annotations.

In contrast to the previous datasets, SciPostLayout focuses on scientific posters.
The layout analysis of scientific posters is more challenging than that of scientific papers because of the variety of fonts and positions of figures and tables.


\subsection{Layout Generation}

Layout generation is a task that arose from the needs of design applications, including magazine covers, UI interfaces, presentation slides, and banner advertising~\cite{deka2017rico,fu2022doc2ppt,https://doi.org/10.1111/cgf.14399,yamaguchi2021canvasvae,zheng2019content}.
Layout generation can be categorized into unconditional generation~\cite{arroyo2021variational,gupta2021layouttransformer,jiang2022coarse}, which generates layouts without any constraints from the user, and conditional generation, which enables the user to create their own desired layout with constraints.
There are also several subcategories within conditional generation, each giving the system different constraints for generation, such element types~\cite{Kikuchi2021,kong2022blt,10.1007/978-3-030-58580-8_29}, element types and sizes~\cite{kong2022blt}, relationships between elements~\cite{Kikuchi2021,10.1007/978-3-030-58580-8_29}, completion~\cite{gupta2021layouttransformer}, and refinement~\cite{rahman2021ruite}.

Early studies explored generative adversarial networks (GANs)~\cite{Kikuchi2021,li2019layoutgan} and variational autoencoders (VAEs)~\cite{arroyo2021variational,jiang2022coarse,yamaguchi2021canvasvae} for layout generation.
Motivated by the success of the Transformer architecture~\cite{vaswani2017attention} in NLP, some studies treated layout generation as a sequence-to-sequence problem~\cite{gupta2021layouttransformer,Jiang_2023_CVPR,kong2022blt,lin2023layoutprompter}.
Moreover, discrete diffusion models have achieved notable results~\cite{chai2023layoutdm,Hui2023UnifyingLG,inoue2023layoutdm,zhang2023layoutdiffusion}.
Datasets including Rico~\cite{deka2017rico}, Magazine~\cite{zheng2019content}, and PubLayNet~\cite{zhong2019publaynet} are typically utilized to evaluate layout generation models.

We evaluate existing models for layout generation in the scientific poster domain.
Generating layouts for scientific posters is challenging, as are other layout generation tasks that require the positioning of figures and tables.


\subsection{Scientific Poster Generation}

As stated earlier, scientific poster generation is a task in which a system generates a poster summarizing the important information in a scientific paper.
Although this task is promising for applications, research on scientific poster generation remains scarce due to its high complexity and multimodality.

Paramita and Khodra~\cite{7803112} treated scientific poster generation as a summarization problem and developed a model to extract essential sentences from papers by classification and fill them into templates to generate posters.
However, the generated posters are composed only of texts, making them unsuitable for efficiently conveying information.

Qiang et al.~\cite{1547260908565-217496438} decomposed poster generation into three subtasks: content extraction, panel attribute inference, and panel layout generation.
For content extraction, their system uses TextRank~\cite{mihalcea-tarau-2004-textrank} to summarize each section of a paper and then manually pick the important figures and tables.
For panel attributes inference, the system uses a Bayesian network trained to learn the size and aspect ratio of each panel in a poster.
For panel layout generation, the system generates panel layouts with a binary tree structure recursively.
Finally, with the inferred data, posters are generated in a LaTeX format.
The authors also constructed a dataset with 85 paper-poster pairs and facilitated evaluations of the model performances by experts on three metrics: readability, informativeness, and aesthetics.
However, the licensing status of the posters is unclear, and the layout annotation is not detailed enough; for example, both figures and tables are simply labeled as \textit{Figure}.

Xu and Wan~\cite{xu2021neural} focused on the content extraction part of poster generation and proposed a model to extract text, figures, and tables from papers simultaneously.
They additionally proposed a three-step framework to generate posters: section filter, content extraction, and poster composition.
They also built a dataset for evaluation, which is currently awaiting public release.

Our SciPostLayout dataset is the first to contain scientific posters and papers with the CC-BY license, which can be used to evaluate scientific poster generation systems.



\section{Dataset}


Previous studies~\cite{1547260908565-217496438,xu2021neural} have built datasets to evaluate poster generation systems for scientific papers.
However, these datasets are either not publicly available or the data licensing is unclear.
In this paper, we construct SciPostLayout, a fully public dataset with all data under the Creative Commons license, which allows unrestricted dissemination, adaptation, and re-use.


First, we downloaded posters in PDF format from F1000Research~\footnote{https://f1000research.com/browse/posters}.
Among these, we retained 7,943 posters under the CC-BY license.
Posters under non-distributable and noncommercial licenses were excluded.
The PDF files were converted into PNG format at DPI=100 for the following annotations.
We excluded posters with file sizes below 200KB as they mainly consisted of text, which is unsuitable for layout analysis.
After this exclusion, 7,855 posters remained.
We investigated the word trends in the titles and found that most of the collected posters were in the biomedical field.

Next, we recruited professional data annotators to manually annotate the document layout of the 7,855 posters.
The layouts of scientific posters are more diverse than the layouts of the papers in PubLayNet~\cite{zhong2019publaynet} because of the variety of fonts, size and position of figures and tables, and typography.
We expanded PubLayNet's five-category annotation standard to nine categories to acquire fine-grained annotations of the layouts.
The annotation criteria are listed in Table~\ref{tab:annotation_criteria}.

\begin{table}[t]
\small
\centering
\begin{tabular}{l@{\hskip 0.5in}l}
\hline
Category & Contents \\
\hline
Title & paper title\\
Author Info & author, author affiliation\\
Section & section title$^a$\\
Text & paragraph$^b$\\
List & nested list$^c$\\
Table & main body of table\\
Figure & main body of figure$^d$\\
Caption & caption of table and figure\\
Unknown & advertising information, logo of affiliation$^e$\\
\hline
\end{tabular}

\raggedright
\vspace*{1ex}
\footnotesize{$^a$}\footnotesize{Only the highest level of sections are annotated as Section. Subsections are annotated as Text because of the small font size.}\\
\footnotesize{$^b$}\footnotesize{When multiple paragraphs are connected, they are annotated as a single object. When a subsection is inserted between paragraphs, a new bounding box is created. The subsection name is not enclosed in a separate bounding box, but is included together with the subsequent paragraph.} Footnotes are ignored during annotation.\\
\footnotesize{$^c$}\footnotesize{Not only bullet points ($\cdot$) but also paragraph numbers (e.g. 1., a.) are treated as lists. It includes reference block.}\\
\footnotesize{$^d$}\footnotesize{When sub-figures exist, the whole figure panel is annotated as a single object.}\\
\footnotesize{$^e$}\footnotesize{Only elements with big areas are annotated.}\\
\vspace*{1ex}
\caption{Annotation criteria of SciPostLayout.}
\label{tab:annotation_criteria}
\end{table}

Subsequently, we manually searched for the papers paired with the posters because there is no information relating to the papers on F1000Research.
We found 100 papers under the CC-BY license and then divided these 100 paper-poster pairs into dev/test sets comprising 50 pairs each.

After the above processes, we obtained our dataset SciPostLayout, which includes train data with 6,855 posters and manual layout annotations as well as dev/test data with 500 posters, 50 papers paired with partial posters, and manual layout annotations.
Table~\ref{tab:statistics} lists the statistics of the dataset.
SciPostLayout is the first scientific poster layout dataset that can be used for evaluating layout analysis and layout generation systems.
Moreover, SciPostLayout is the first dataset that contains scientific paper-poster pairs with the CC-BY license, which can be utilized for evaluating scientific poster generation systems.

\begin{table*}[t]
\centering
\small
\resizebox{0.95\linewidth}{!}{
\begin{tabular}{l|rrrrrrrrr|r}
\hline
Split & Title & Author Info & Section & Text & List & Table & Figure & Caption & Unknown & Total \\
\hline
Train & 6,847 & 6,709 & 36,083 & 46,297 & 20,723 & 4,916 & 33,665 & 13,065 & 8 & 168,313 \\
Dev & 497 & 492 & 2,609 & 3,215 & 1,489 & 401 & 2,421 & 1,015 & 2 & 12,141 \\
Test & 498 & 494 & 2,646 & 3,199 & 1,583 & 328 & 2,484 & 1,010 & 1 & 12,243 \\
\hline
\end{tabular}
}
\caption{Statistics of train, dev, and test data in SciPostLayout.}
\label{tab:statistics}
\end{table*}


\section{Experiments}

We conducted three experiments using the collected dataset: layout analysis, layout generation, and paper-to-layout. 
We compared existing models for each experiment.


\subsection{Layout Analysis}
\label{sec:layout_analysis}

We used LayoutLMv3~\cite{huang2022layoutlmv3} and DiT~\cite{li2022dit} for the layout analysis task.
We started from the base size checkpoints with Cascade R-CNN detectors and fine-tuned them on SciPostLayout train data for both models.
The checkpoint with the highest performances on the dev set was used for evaluation.

We measured the performance using the mean average precision (mAP) @ intersection over union (IoU) [0.50:0.95] of bounding boxes, the results of which are reported in Table~\ref{tab:result:layout_analysis}. 
The \textit{Unknown} category was omitted due to an insufficient number of elements.
LayoutLMv3 outperformed DiT in all categories.
In addition, both models showed high performances in the \textit{Title} and \textit{Author Info} categories.
We attribute this result to the regularity of the \textit{Title} and \textit{Author Info} blocks, since they are always at the top of the posters and there is usually only one of each block per poster.
However, compared to the results on PubLayNet~\cite{huang2022layoutlmv3,li2022dit}, in which mAP@IoU is over 90, both models showed a performance drop, indicating the complexity of our dataset.

\begin{table*}[t!]
\centering
\small
\resizebox{0.95\linewidth}{!}{
\begin{tabular}{l|rrrrrrrr|r}
\hline
Model & Title & Author Info & Section & Text & List & Table & Figure & Caption & Overall \\
\hline
LayoutLMv3 & \textbf{88.32} & \textbf{85.79} & \textbf{71.98} & \textbf{71.33} & \textbf{73.71} & \textbf{74.41} & \textbf{67.60} & \textbf{69.24} & \textbf{66.93} \\
DiT & 85.23 & 79.89 & 70.64 & 67.66 & 69.93 & 74.16 & 67.12 & 59.27 & 63.77 \\
\hline
\end{tabular}
}
\caption{Layout analysis performance (mAP@IoU[0.50:0.95]) on SciPostLayout test set. \textbf{Bold} numbers indicate highest performances.}
\label{tab:result:layout_analysis}
\end{table*}


\subsection{Layout Generation}
\label{sec:layout_generation}

We conducted layout generation experiments with various settings for the information to be input into the model~\cite{Jiang_2023_CVPR}.
Generation conditioned on types (Gen-T) aims to generate layouts from the number of element types (categories).
Generation conditioned on types and sizes (Gen-TS) aims to generate layouts from the number and size of element types.
Generation conditioned on relationships (Gen-R) aims to generate layouts from the number of element types and position relationships between the elements.
Completion means generating a complete layout from a part of the layout.
Refinement means generating a new layout from a layout that needs improvement.

We used LayoutDM~\cite{inoue2023layoutdm}, LayoutFormer++~\cite{Jiang_2023_CVPR}, and LayoutPrompter~\cite{lin2023layoutprompter} for the layout generation task.
Unlike the layout analysis models, the layout generation models were trained from randomly initialized parameters.
We did not train LayoutPrompter because it uses GPT-4\footnote{gpt-4-1106-preview} via API.

We evaluated the model performances using maximum IoU (mIoU), Alignment, Overlap, and Fr\'{e}chet Inception Distance (FID).
mIoU is a measure of the highest IoU between a generated layout and a real layout~\cite{Kikuchi2021,lin2023layoutprompter}.
Alignment indicates how well the elements in a layout are aligned with each other~\cite{9106863,lin2023layoutprompter}.
Overlap is the overlapping area between two arbitrary elements in a layout~\cite{9106863,lin2023layoutprompter}.
FID measures how similar the distribution of the generated layouts is to that of real layouts~\cite{Kikuchi2021,lin2023layoutprompter}.
A higher mIoU value means higher performance; for the other metrics, a lower value means higher performance.
Note that both mIoU and FID are evaluation metrics based on similarity to the real layouts, but mIoU is calculated from the intersection between layouts, while FID is an embedding-based evaluation metric.
Thus, the hierarchical order of mIoU and FID performances across models is not always consistent.

\begin{figure}[t!]
\centering
\includegraphics[width=0.8\linewidth]{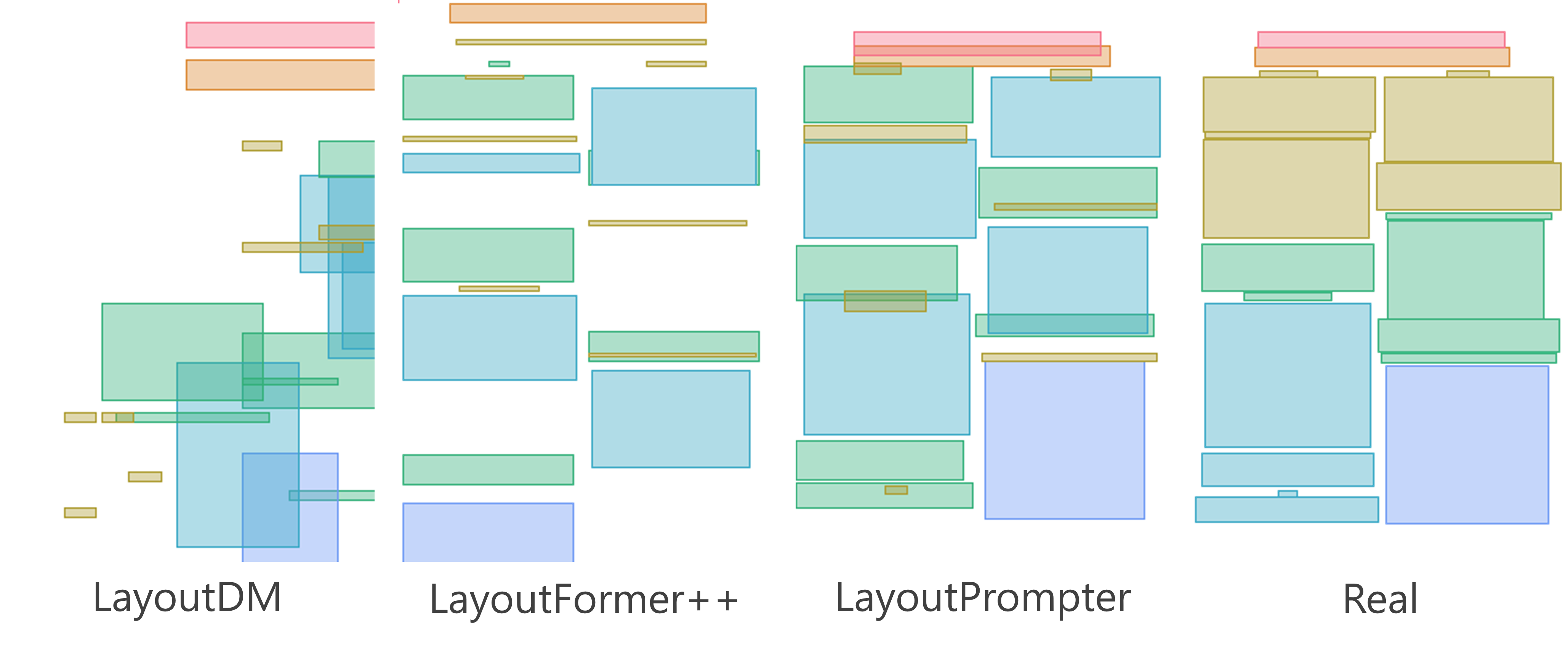}
\vspace{-3mm}
\caption{Examples of generated layouts and the real layout in the Refinement setting. Examples of the other settings are provided in Appendix A.}
\label{fig:refinement_example}
\end{figure}

The results are shown in Table~\ref{tab:result:layout_generation}.
mIoU was low for all models and was less than half that in PubLayNet~\cite{inoue2023layoutdm,Jiang_2023_CVPR,lin2023layoutprompter}.
In contrast, all models performed effectively on Alignment, indicating that they can generate aligned layouts.
LayoutPrompter was the most effective for Overlap, indicating that it generates layouts with the least overlap.
LayoutDM was the most effective in terms of FID, indicating that it can generate layouts most similar to real layout distribution on each setting.
Comparing the FID and Overlap of LayoutDM and LayoutPrompter shows that LayoutDM tends to generate layouts that are similar to the real layout distribution but with more overlap, while LayoutPrompter generates layouts with less overlap but that are farther from the real layout distribution.
In the Refinement setting, LayoutPrompter outperformed the other models, indicating that it can generate layouts similar to real layouts from noisy layouts.
Figure~\ref{fig:refinement_example} shows examples of generated layouts of each model and the real layout in the Refinement setting.
LayoutPrompter generated the most similar layout to the real layout, while LayoutDM and LayoutFormer++ generated layouts with overlap that were not similar to the real layout.

\begin{table*}[t!]
\centering
\small
\begin{tabular}{l|l|rrrr}
\hline
Task & Model & mIoU$\uparrow$ & Alignment$\downarrow$ & Overlap$\downarrow$ & FID$\downarrow$ \\
\hline
      & LayoutDM & 0.073 & 0.002 & 0.524 & \textbf{1.524} \\
Gen-T & LayoutFormer++ & 0.079 & 0.002 & 0.242 & 26.999 \\
      & LayoutPrompter & \textbf{0.084} & \textbf{0.000} & \textbf{0.231} & 9.470 \\
\hline
       & LayoutDM & \textbf{0.113} & 0.002 & 0.564 & \textbf{1.419} \\
Gen-TS & LayoutFormer++ & 0.089 & 0.003 & 0.245 & 27.372 \\
       & LayoutPrompter & 0.102 & \textbf{0.000} & \textbf{0.182} & 8.605 \\
\hline
      & LayoutDM & \textbf{0.075} & 0.001 & 0.571 & \textbf{5.556} \\
Gen-R & LayoutFormer++ & 0.066 & 0.002 & 0.553 & 20.296 \\
      & LayoutPrompter & 0.069 & \textbf{0.000} & \textbf{0.317} & 6.282 \\
\hline
           & LayoutDM & -- & 0.001 & 0.610 & \textbf{2.490} \\
Completion & LayoutFormer++ & -- & 0.001 & 0.250 & 5.840 \\
           & LayoutPrompter & -- & \textbf{0.000} & \textbf{0.011} & 10.874 \\
\hline
           & LayoutDM & 0.119 & 0.002 & 0.531 & 1.196 \\
Refinement & LayoutFormer++ & 0.301 & 0.002 & 0.289 & 10.955 \\
           & LayoutPrompter & \textbf{0.552} & \textbf{0.001} & \textbf{0.112} & \textbf{0.152} \\
\hline
\end{tabular}
\caption{Layout generation performance on SciPostLayout test set. \textbf{Bold} numbers indicate the highest performances. $\uparrow$ indicates larger values are ideal, $\downarrow$ indicates smaller values are ideal. mIoU is omitted from the Completion setting because the number of elements of generated layouts differs from that of real layouts.}
\label{tab:result:layout_generation}
\end{table*}

\subsection{Paper-to-Layout}

\begin{figure}[t!]
\centering
\includegraphics[width=0.8\linewidth]{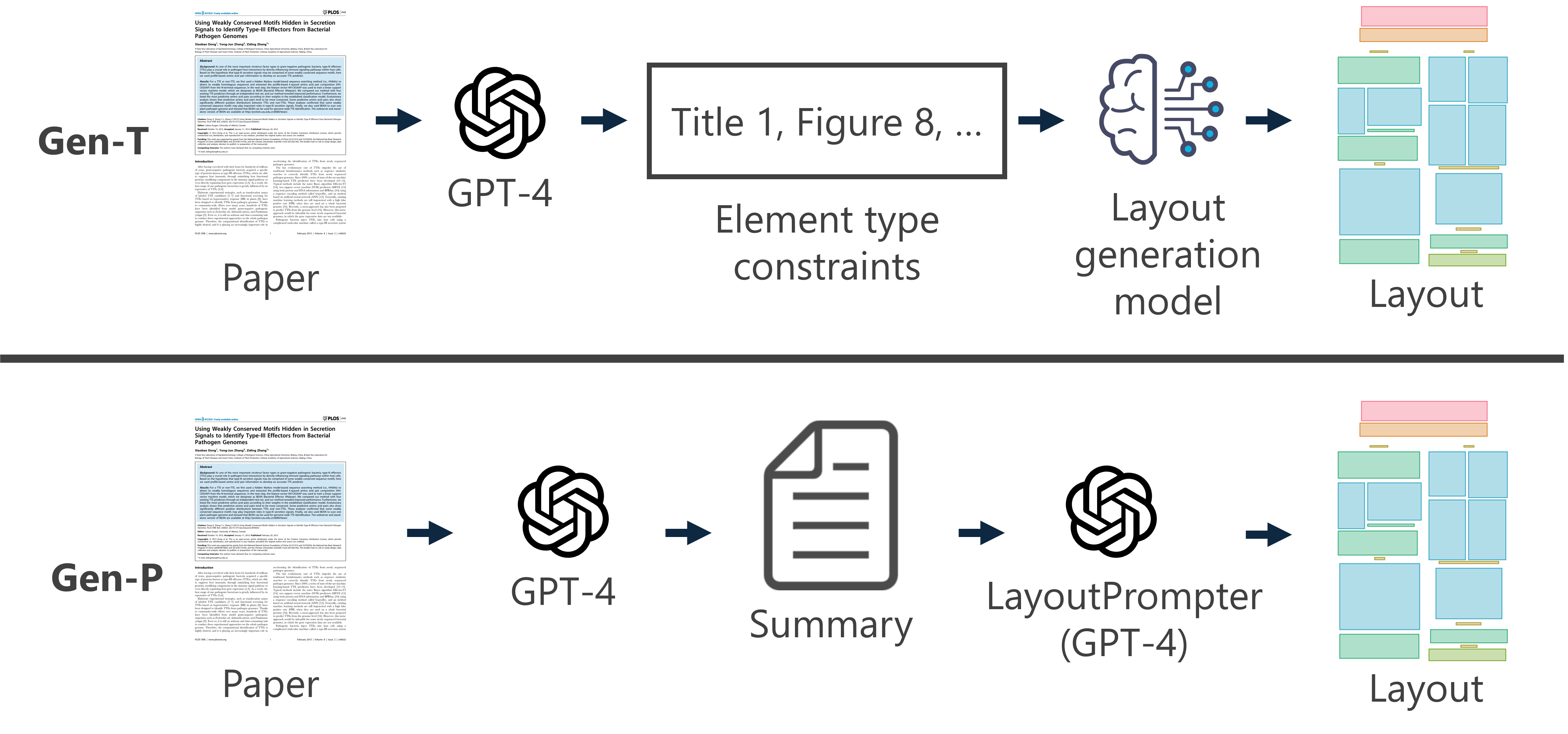}
\vspace{-3mm}
\caption{Gen-T and Gen-P settings of Paper-to-Layout.}
\label{fig:gent_and_genp}
\end{figure}

In Section~\ref{sec:layout_generation}, we conducted experiments on layout generation from real constraints.
To develop a system that automatically generates a poster from a scientific paper, we need to implement a model that extracts constraints or generates a layout from a paper.
Thus, we implement and compare models to generate layouts from scientific papers in two settings: Gen-T and Gen-P.
Figure~\ref{fig:gent_and_genp} illustrates the settings of Gen-T and Gen-P.
In the Gen-T setting, GPT-4 first extracts element type constraints from a scientific paper.
Then, a layout generation model generates a layout from the extracted constraints, the same as the Gen-T setting in Section~\ref{sec:layout_generation}.
In the Gen-P setting, GPT-4 first generates a summary within 1,000 words from a scientific paper.
Then, LayoutPrompter with GPT-4 generates a layout from the paper summary.
LayoutPrompter does not directly generate a layout from a paper because the entire paper is too long for document retrieval when it searches for examples similar to the target paper for layout generation, which are given as few-shot examples.
We used 50 paper-poster pairs of dev set as few-shot examples in the Gen-P setting.
Different from Section~\ref{sec:layout_analysis} and ~\ref{sec:layout_generation}, we used only 50 paper-poster pairs of test set to evaluate the model performances.
In both settings, Nougat~\cite{blecher2023nougat} was utilized to extract texts from a paper in PDF format.
The prompts to extract element type constrains or generate a paper summary are shown in Appendix B.

We evaluated the accuracy of extracting element type constraints in the Gen-T setting based on mean absolute error (MAE).
MAE is calculated as the mean of the absolute differences between the number of elements predicted by GPT-4 and the actual number of elements in the poster layout.
For instance, if GPT-4 predicts 4 \textit{Figures} while the actual poster has 5, the absolute error for this element type would be |4-5| = 1.
We performed extraction three times and calculated the mean value.
Table~\ref{tab:mae_results} shows the results.
For \textit{Title} and \textit{Author Info}, the generated number of elements is close to the real constraints.
For \textit{Text} and \textit{Figure}, GPT-4 shows large error values because the number of these elements is larger than the number of other elements and there is also a large range of values, as shown in Table~\ref{tab:statistics}.
However, the overall MAE is less than 2, indicating that GPT-4 can generate constraints close to the real number of elements.

\begin{table*}[t]
\centering
\small
\begin{tabular}{l|rrrrrrrr|r}
\hline
Model & Title & Author Info & Section & Text & List & Table & Figure & Caption & Overall \\
\hline
GPT-4 & 0.06 & 0.04 & 1.90 & 4.9 & 2.31 & 0.57 & 3.66 & 2.43 & 1.98 \\
\hline
\end{tabular}
\caption{MAE results of element type extraction by GPT-4.}
\label{tab:mae_results}
\end{table*}

\begin{figure}[t!]
\centering
\includegraphics[width=0.6\linewidth]{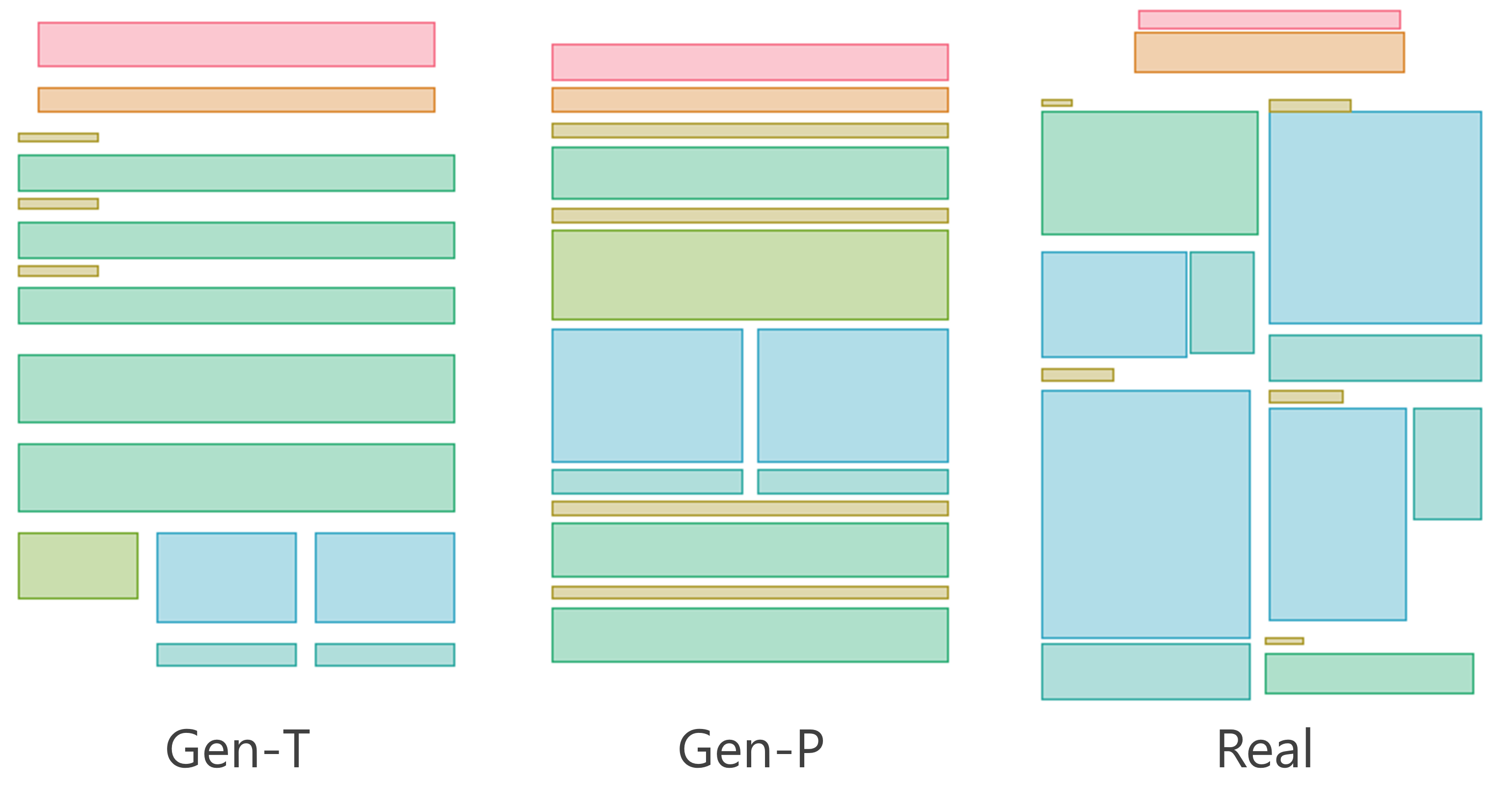}
\vspace{-3mm}
\caption{Example of LayoutPrompter generated layout and real layout in the Gen-T and Gen-P setting from scientific paper.}
\label{fig:gen_example_paper2poster}
\end{figure}

We conducted experiments on poster layout generation from scientific papers in the Gen-T and Gen-P settings, which are reported in Table~\ref{tab:result:paper2layout}.
The number of each element of the extracted constraints in the Gen-T setting was given to the models as the mean value rounded down to the nearest whole number.
All models performed effectively on Alignment, indicating that they can generate aligned layouts in the same way as layout generation from real constraints in Section~\ref{sec:layout_generation}.
LayoutPrompter in the Gen-P setting was the most effective for Overlap, indicating that it generates layouts with the least overlap.
Comparing the FID values in the Gen-T setting to the values in the Gen-T setting of Table~\ref{tab:result:layout_generation}, we find that the FID values of all models slightly increase, indicating that generating layouts similar to real layouts from extracted constraints is more difficult than generating layouts from real constraints.
Figure~\ref{fig:gen_example_paper2poster} shows examples of layouts generated by LayoutPrompter in the Gen-T and Gen-P settings.
Although the generated layouts are not similar to the real layout, LayoutPrompter could generate aligned layouts with less overlap.
The experimental results in this section demonstrate that although the current LLM has difficulty generating layouts similar to real layouts from scientific papers, it can generate high-quality layouts, indicating that LLM has potential as a poster generation system from scientific papers.

\begin{table*}[t!]
\centering
\small
\begin{tabular}{l|l|rrr}
\hline
Task & Model & Alignment$\downarrow$ & Overlap$\downarrow$ & FID$\downarrow$ \\
\hline
      & LayoutDM & 0.001 & 0.542 & \textbf{9.073} \\
Gen-T & LayoutFormer++ & 0.001 & 0.617 & 56.396 \\
      & LayoutPrompter & \textbf{0.000} & 0.256 & 19.892 \\
\hline
Gen-P & LayoutPrompter & \textbf{0.000} & \textbf{0.022} & 17.790 \\
\hline
\end{tabular}
\caption{Paper-to-Layout generation performance on SciPostLayout test set. \textbf{Bold} numbers indicate the highest performances. $\downarrow$ indicates smaller values are ideal. mIoU is omitted because the number of elements of generated layouts differs from that of real layouts.}
\label{tab:result:paper2layout}
\end{table*}

\section{Conclusion}

In this paper, we built a new dataset called SciPostLayout, which consists of 7,855 scientific posters and 100 scientific papers downloaded from a website.
All posters in SciPostLayout are manually annotated with categories of layout elements (such as titles and figures) to be used in layout analysis and generation tasks.
In addition, SciPostLayout is available for commercial research because all of the posters and papers are under the CC-BY license.
We conducted layout analysis and generation experiments to evaluate the performances of existing models on SciPostLayout.
For layout analysis, although some elements could be recognized with high accuracy, we found that layout analysis on SciPostLayout was more difficult than on a scientific paper dataset.
For layout generation, although existing models could generate aligned layouts, we found it was difficult to generate layouts that are similar to real layouts.
In addition, we implemented and evaluated GPT-4-based models to generate poster layouts from scientific papers.
The experimental results proved that LLM has potential as a scientific poster generation system.
Our future work will involve developing a model to improve layout analysis and generation.
In addition, we will investigate improved methods for generating poster layouts and extracting contents from scientific papers.

\section*{Acknowledgements}
This work was supported by JST Moonshot R\&D Program, Grant Number JPMJMS2236.
We would like to thank Naoto Inoue, Ph. D. of CyberAgent Inc. AILab for useful comments about layout generation models.

\bibliography{egbib}
\end{document}


\maketitle

\begin{appendices}
\section{Additional Examples of Layout Generation}

\begin{figure}[t!]
\centering
\includegraphics[width=0.9\linewidth]{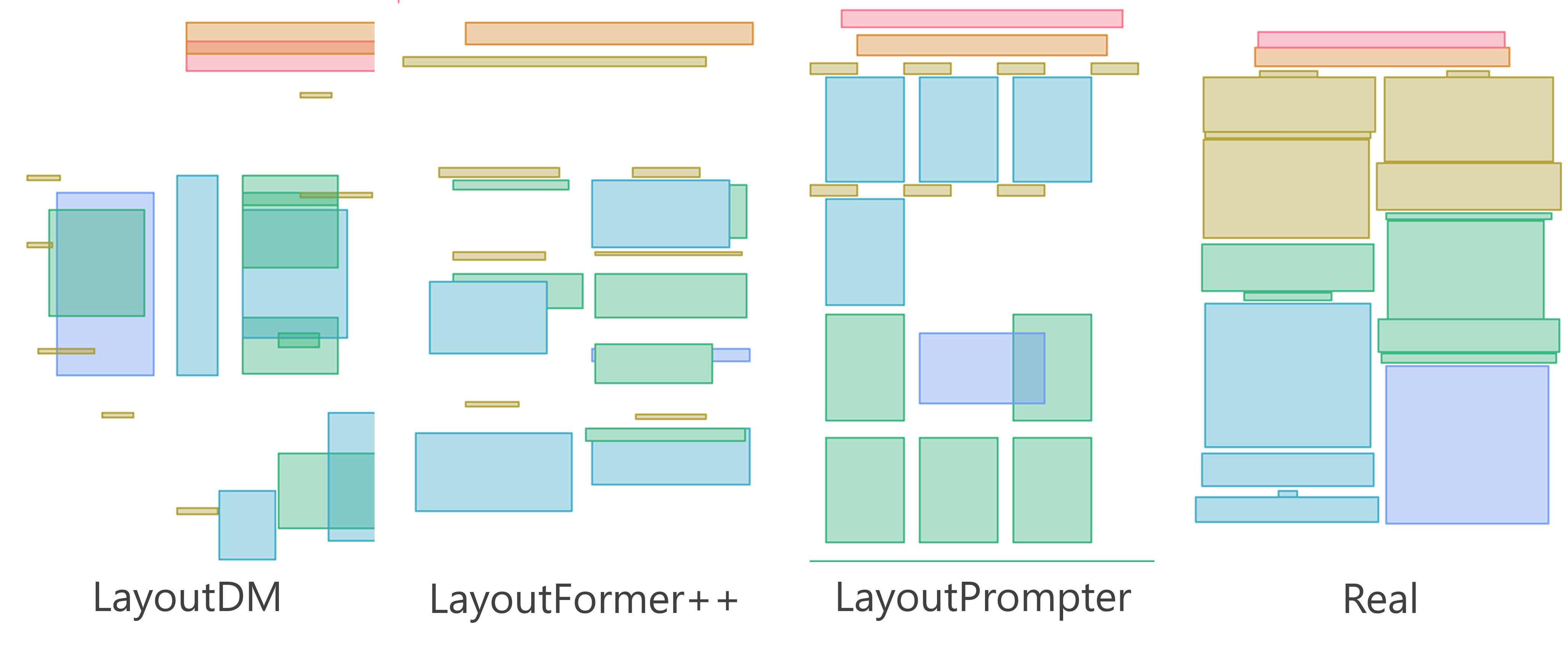}
\caption{Examples of generated layouts and the real layout in the Gen-T setting.}
\label{fig:gent_example}
\end{figure}

\begin{figure}[t!]
\centering
\includegraphics[width=0.9\linewidth]{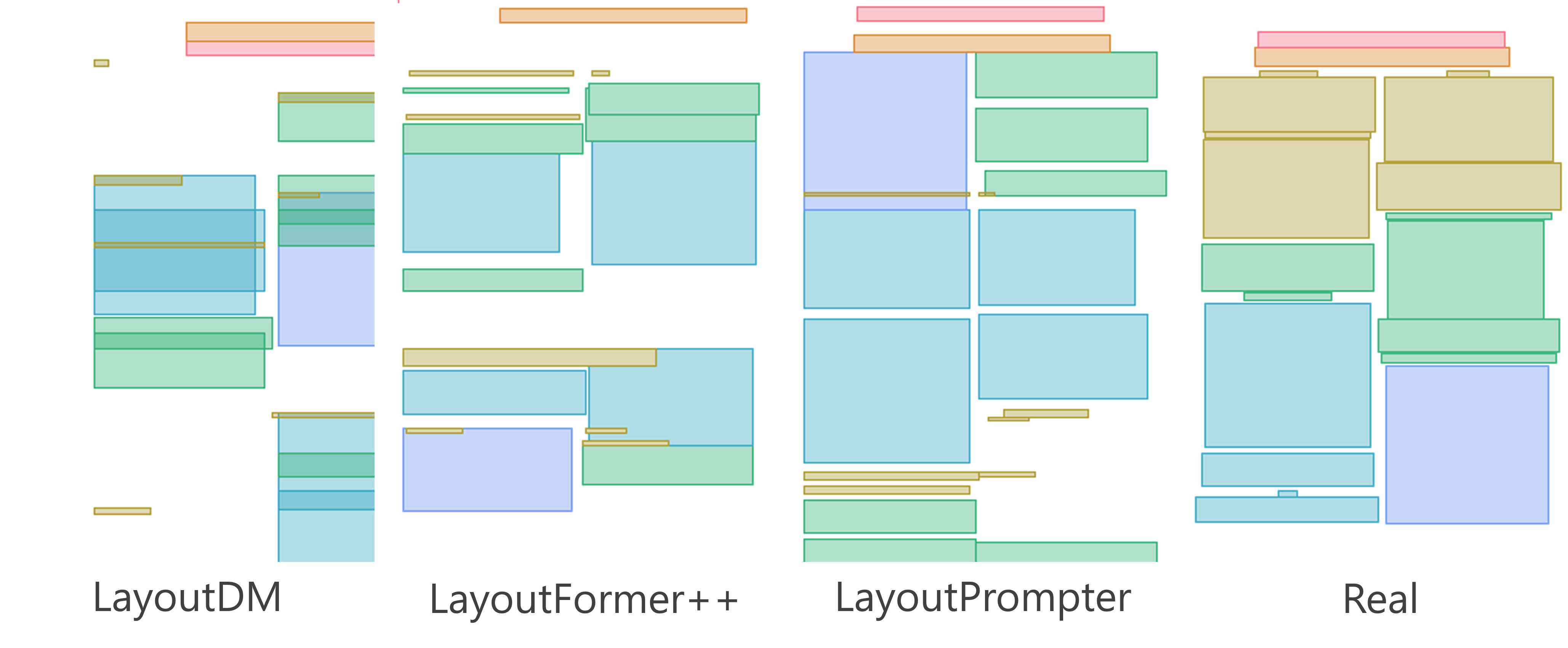}
\caption{Examples of generated layouts and the real layout in the Gen-TS setting.}
\label{fig:gents_example}
\end{figure}

\begin{figure}[t!]
\centering
\includegraphics[width=0.9\linewidth]{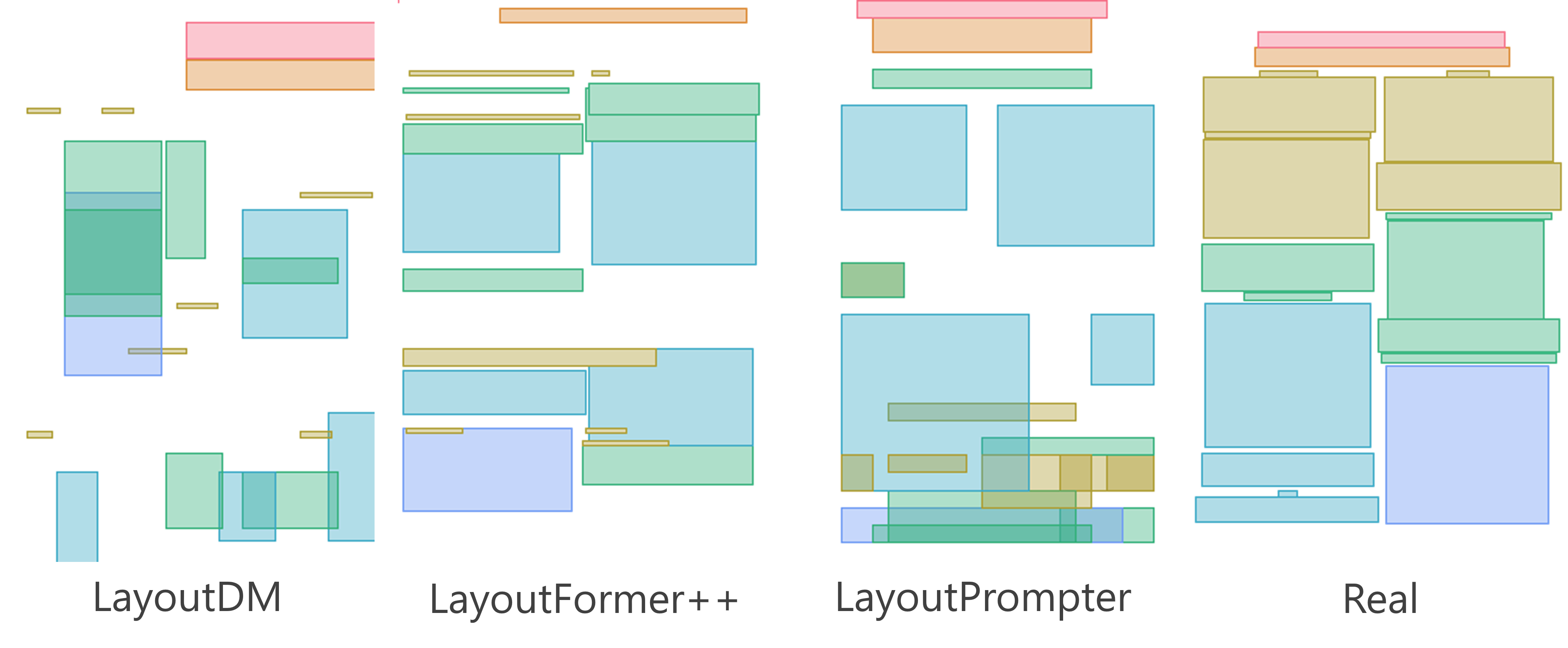}
\caption{Examples of generated layouts and the real layout in the Gen-R setting.}
\label{fig:genr_example}
\end{figure}

\begin{figure}[t!]
\centering
\includegraphics[width=0.9\linewidth]{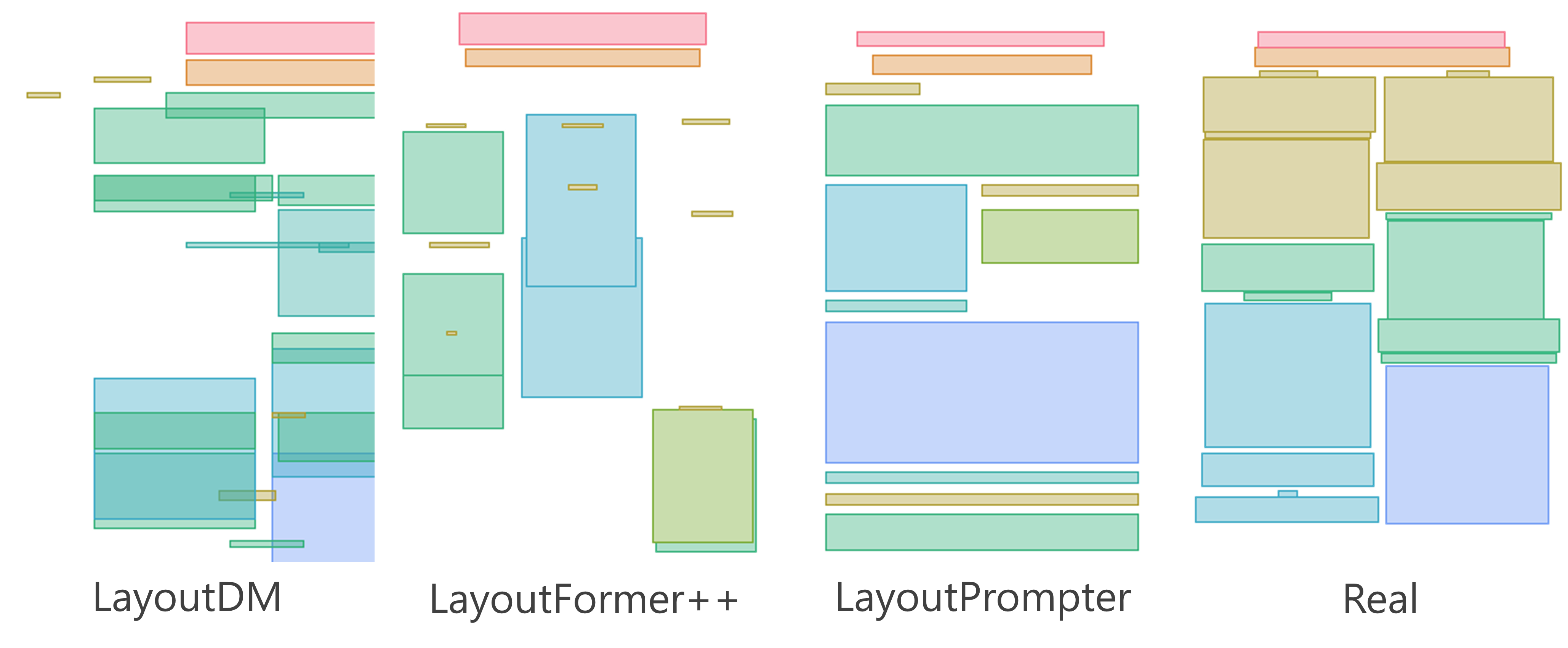}
\caption{Examples of generated layouts and the real layout in the Completion setting.}
\label{fig:completion_example}
\end{figure}

\begin{figure}[t!]
\centering
\includegraphics[width=0.9\linewidth]{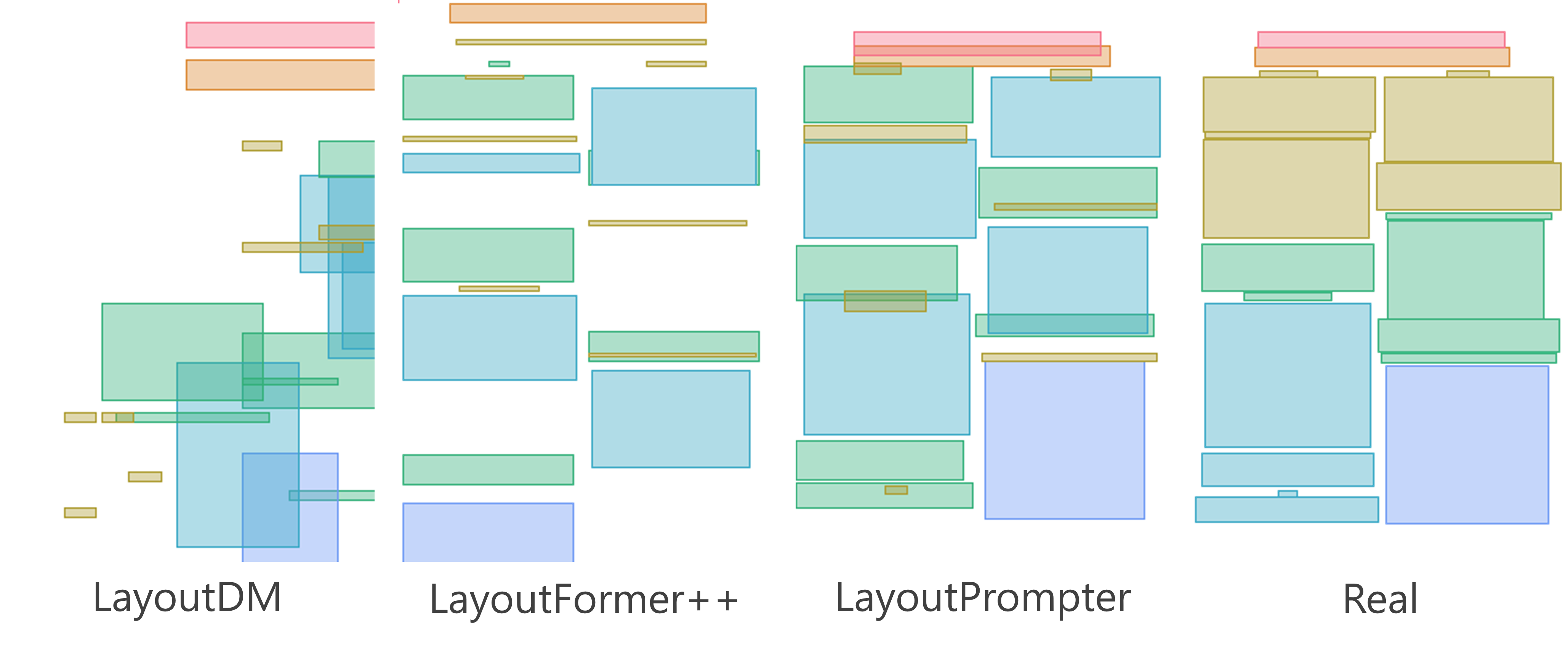}
\caption{Examples of generated layouts and the real layout in the Refinement setting.}
\label{fig:refinement_example}
\end{figure}

Figure~\ref{fig:gent_example}-\ref{fig:refinement_example} shows example layouts which the models generated in layout generation experiment of Section 4.2.
As shown in Section 4.2, LayoutPrompter tends to generate aligned layouts with less overlap.
LayouDM generated layouts with overlap that are not similar to the real layouts.
In Section 4.2, the FID values of LayoutDM tend to be lower than the values of other models, but larger than the FID value of LayoutPrompter in the Refinement setting, which was able to generate similar layouts to the real layouts.
In other words, LayoutDM can generate layouts that similar to the real layout distribution than other models, but it cannot generate layouts that can be judged similar to the real layouts in subjective evaluation.

\clearpage

\clearpage
\section{Prompts for Paper-to-Layout}

Listing~\ref{list:prompt_rule} is the prompt to extract element type constrains from a scientific paper in the Gen-T setting in Section 4.3.

\begin{lstlisting}[caption=Prompt to extract elemet type constrains from a scientific paper, label=list:prompt_rule]
Task: Analyzing and Extracting Layout Generation Constraints for Scientific Poster

You are provided with a parsed text file extracted from a scientific paper in PDF format. Your goal is to analyze the content and determine the layout generation constraints for a scientific poster. The poster should be structured into the following eight categories:

1.Title: The title of the paper. Typically, there's only one block in a poster.
2.Author Info: Authors' names and affiliations. Typically, there's only one block in a poster.
3.Section: Section names. Include only important sections.
4.Text: Paragraphs. Merge paragraphs describing similar content into one block.
5.List: Itemization, including experimental procedure, enumeration of conditions, reference list, etc.
6.Table: Tables.
7.Figure: Figures.
8.Caption: Captions of tables and figures. The number of captions should match the combined total of 6. Table and 7. Figure.

The goal is to determine the number of elements for each category that should be included in the generated poster. Consider that not all elements from the paper need to be included in the poster; prioritize the most important parts. Provide a detailed breakdown of the number of elements for each category. For example, if the paper includes 1 Title, 1 Author Info, 5 Section, 10 Text, 2 List, 1 Table, 2 Figure and 3 Caption blocks, specify these quantities for efficient poster generation.

Please return the results in JSON format, don't say anything else.

Below is the text file:
\end{lstlisting}

\clearpage
Listing~\ref{list:prompt_summary} is the prompt to generate a summary of a scientific paper in the Gen-P setting in Section 4.3.

\begin{lstlisting}[caption=Prompt to generate a summary of a scientific paper, label=list:prompt_summary]
Please summarize the following paper within 1000 words.
The summary does not need to include all elements of the paper, but should prioritize important elements such as proposed methods and main experimental results.
The summary should include the title and the author names of the paper.
Use the same wording as in the paper's abstract.
DO NOT generate redundant messages.
\end{lstlisting}

\end{appendices}